\documentclass[letterpaper, 10 pt, conference]{ieeeconf}  
\IEEEoverridecommandlockouts                              
\usepackage{amsmath,amsfonts}
\usepackage{algorithmic}
\usepackage{algorithm}
\usepackage{array}
\usepackage{textcomp}
\usepackage{stfloats}
\usepackage{url}
\usepackage{verbatim}
\usepackage{graphicx}
\usepackage{subcaption}
\usepackage{cite}
\usepackage{hyperref}
\usepackage{tabularx}
\usepackage{booktabs}
\usepackage{comment}
\usepackage[section]{placeins}

\title{\LARGE \bf
SteinPnP: Non-Parametric Uncertainty Quantification \\ in the Perspective-N-Point Problem
}

\author{Baptiste Lardinoit, Jongseok Lee, Andres Yubero, Seong Hun Lee, Javier Civera, and Rudolph Triebel}

\begin{document}


\maketitle
\thispagestyle{empty}
\pagestyle{empty}

\begin{abstract}
We propose SteinPnP, a probabilistic Perspective-N-Point (PnP) solver that models the full 6-DoF camera pose posterior using a non-parametric set of pose particles. By leveraging Stein Variational Inference (SVI), SteinPnP combines the expressivity of particle-based representations with the efficiency of optimization-based solvers. However, a naive application of SVI struggles in the PnP problem, often missing alternative pose hypotheses or assigning incorrect probability weights to them. To address these limitations, we propose practical remedies that discover new potential poses, identify distinct hypothesis clusters, and balance their relative probability weights to capture the true underlying distribution. Our experiments across synthetic and real-world benchmarks demonstrate that SteinPnP accurately captures complex multimodal uncertainties while maintaining competitive accuracy.  
\end{abstract}
\section{Introduction}

\begin{figure}[t]
    \centering
    \hspace*{-7mm}
    \includegraphics[width=0.9\linewidth]{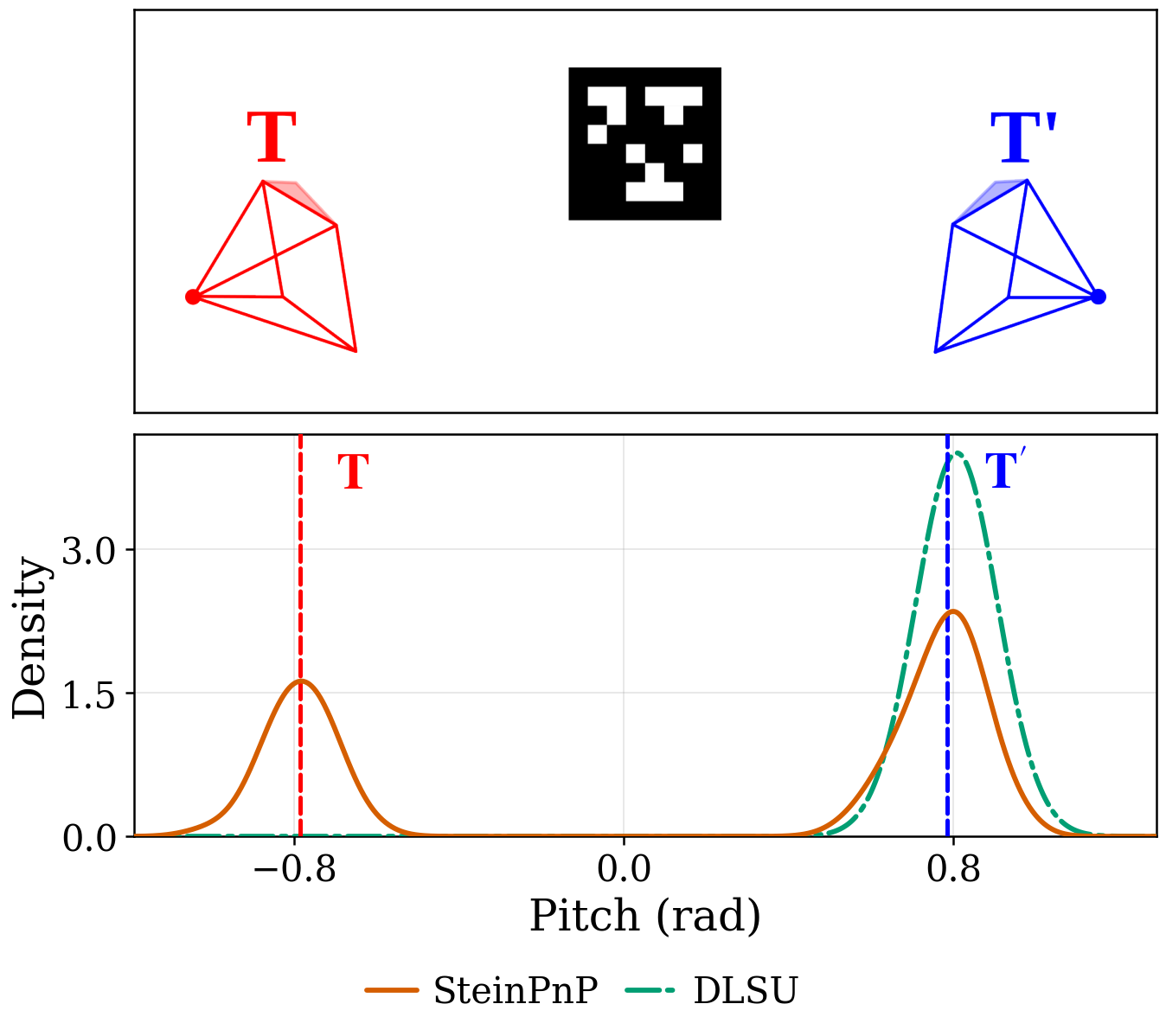}
    \caption{Top: observation of an AprilTag from the true pose \(\mathrm{T}\) and the ambiguous pose \(\mathrm{T}^{\prime}\). Since the tag is planar, its projections from \(\mathrm{T}\) and \(\mathrm{T}^{\prime}\) are virtually identical, making pose estimation inherently ambiguous. Bottom: kernel density estimates of the marginal pitch density for SteinPnP and DLSU (a Gaussian method). DLSU predicts the ambiguous pose with overconfidence, whereas SteinPnP correctly captures the uncertainty, assigning probabilities to both pose hypotheses. Only pitch is shown for illustration but SteinPnP estimates the full 6-DoF pose distribution.}
    \vspace*{-6mm}
\label{fig:teaser}
\end{figure}

Estimating the 6-DoF pose of a camera, \emph{i.e.}, its position and orientation, from visual measurements is a fundamental problem in robotic perception. In robot navigation, pose estimation forms the core of state estimation and mapping, with its applications including visual odometry \cite{fraundorfer2012visual}, 3D reconstruction \cite{agarwal2011building}, and simultaneous localization and mapping (SLAM) \cite{grisetti2010tutorial}. In robotic manipulation, estimating 6-DoF object poses is likewise essential for capabilities such as object localization \cite{chen2022epro}, rigid object pose tracking \cite{stoiber2022iterative}, and grasp prediction \cite{chitta2012moveit}. A central formulation underlying these tasks is the Perspective-N-Point (PnP) problem, which estimates the absolute 6-DoF pose of a calibrated camera from a set of $N$ known 2D-to-3D point correspondences. Over the past decades, numerous solutions to the PnP problem have been proposed to enhance accuracy and computational efficiency.

Although existing PnP solvers focus primarily on accuracy and cost, quantifying the uncertainty over the estimated 6-DoF pose is equally relevant in many applications, including robotics. In practice, sensor noise and visual ambiguities, from symmetric objects or distant keypoints, adversely affect the PnP solver's performance. Characterizing this uncertainty can help robots to prevent failures by avoiding unsafe actions. Moreover, downstream tasks generally depend on uncertainty estimates. In SLAM \cite{grisetti2010tutorial}, for instance, visual pose estimates are often fused with IMU or wheel encoder readings according to their covariances. Overconfident PnP estimates may cause the system to overweight noisy visual measurements, potentially leading to inconsistent maps or localization failures.

Existing probabilistic PnP methods can be broadly categorized based on their input noise assumptions: those considering only 2D feature uncertainty (\emph{e.g.}, CEPnP \cite{ferraz2014leveraging}, MLPnP \cite{mlpnp}), and those modeling both 2D feature and 3D coordinate uncertainties (\emph{e.g.}, EPnP(L)U \cite{vakhitov2021uncertainty}). Generally, these methods represent pose uncertainty using Gaussian distributions. While efficient, Gaussian assumptions fail when visual data admits multiple plausible poses. For instance, symmetric objects, planar keypoints, or distant feature points often produce multiple distinct pose hypotheses that a Gaussian cannot represent. Fig.~\ref{fig:teaser} illustrates such an example based on the widely used AprilTag detectors. Furthermore, existing probabilistic PnP approaches \cite{ferraz2014leveraging, mlpnp, vakhitov2021uncertainty, chen2022epro} use input uncertainties primarily to improve pose estimation accuracy, leaving uncertainty quantification itself statistically unevaluated.

To address these challenges, we propose SteinPnP, a probabilistic algorithm that models PnP pose uncertainty using a non-parametric representation, \emph{i.e.}, a set of pose particles that capture the underlying probability distribution. While Markov Chain Monte Carlo (MCMC) methods are the gold standard for non-parametric uncertainty quantification, they are inevitably slow to converge, and thus not well suited for robotic applications. SteinPnP leverages SVI \cite{liu2016stein}, which combines the flexible non-parametric representation of MCMC with the computational efficiency of optimization-based variational inference. While SVI has been successfully applied to 3D point cloud registration (\emph{e.g.}, Stein ICP \cite{maken2021stein, ma2025svn}), we extend this probabilistic method to 6-DoF pose uncertainty quantification within the PnP problem.

Specifically, we formulate the PnP problem as an approximate Bayesian inference task and design a new PnP solver using a variant of SVI called Stein Variational Newton (SVN) \cite{detommaso2018stein}. In SteinPnP, pose particles initialized from a prior distribution are iteratively driven by SVN toward the target probability distribution. We detail our design choices including the explicit derivations of its score function, Fisher information matrix as a Hessian surrogate, and curvature-scaled kernel design. Importantly, we observe that direct application of SVN suffers from mode collapse and incorrect mass allocation in multimodal PnP settings. To resolve these limitations, we introduce algorithmic remedies featuring randomized initialization, tree-based mode clustering, and importance-sampled mode mass refinement. Finally, we establish a standardized evaluation protocol for PnP uncertainty quantification, demonstrating through extensive synthetic and real-world experiments that non-parametric particles capture visual ambiguities where traditional Gaussian solvers fail.

\textbf{Contributions.} Our key contributions are: (a) SteinPnP, a non-parametric PnP solver modeling 6-DoF camera pose uncertainty via a set of particles (Sec.~\ref{sec:stein_pnp}); (b) identification of mode collapse and incorrect mass allocation issues when applying Stein methods to the PnP problem, and the proposal of practical algorithmic remedies for them (Sec.~\ref{sec:stein_pnp_practice}); (c) an evaluation protocol designed for benchmarking 6-DoF PnP uncertainty quantification (Sec.~\ref{sec:exp:setup}); and (d) empirical validation showing non-parametric particles capture pose ambiguities and failure modes where unimodal Gaussian solvers fail (Sec.~\ref{sec:experiments}).

\section{Related Work}

\textbf{Uncertainty quantification in PnP.} Existing probabilistic PnP methods can be categorized by how they incorporate feature noise into optimization and whether they propagate those uncertainties to the output pose. Initial approaches focused explicitly on 2D feature detection noise. CEPnP \cite{ferraz2014leveraging} models feature noise as anisotropic Gaussians within an unconstrained Sampson error formulation that down-weights noisy keypoints. Similarly, MLPnP \cite{mlpnp} handles 2D feature uncertainties by projecting them into tangent spaces of 3D bearing vectors, and finding a corresponding maximum likelihood formulation. Vakhitov et al. \cite{vakhitov2021uncertainty} broadened input noise models by incorporating both 2D detection noise and 3D point/line uncertainties directly into algebraic residuals, using an iterative reweighting scheme to improve pose accuracy. While other classical solvers use feature noise solely to weight optimization, MLPnP propagates and computes an output pose covariance using first-order linearization.

In the context of 6-DoF object pose estimation, probabilistic PnP methods have been recently integrated with deep learning. Richter-Klug and Frese \cite{richter2019towards} proposed an architecture in which a CNN predicts the coordinates of objects per-pixel and the uncertainties in the image-space. These are fused via a generalized PnP algorithm to produce a pose with a covariance matrix, using a learned correction factor to account for pixel-wise correlation. EPro-PnP \cite{chen2022epro} is another notable example in which a probabilistic PnP layer is devised that outputs a continuous pose distribution. The key idea therein is to exploit the differentiability of the continuous pose distribution in order to learn the object poses end-to-end, including the 2D-3D point correspondences.

These methods demonstrate the relevance of probabilistic approaches in PnP by establishing several non-trivial formulations. However, methods such as MLPnP and Richter-Klug et al. are constrained by their reliance on linearized error propagation, which assumes a unimodal Gaussian posterior. Consequently, they cannot accurately represent the complex, multimodal uncertainty arising from ambiguous PnP scenarios. Furthermore, while EPro-PnP leverages Adaptive Multiple Importance Sampling (AMIS) \cite{cornuet2012adaptive} to capture non-Gaussian pose uncertainty, its scope is orthogonal to ours. EPro-PnP uses its probabilistic layer primarily as a differentiable loss for end-to-end network training, rather than as a standalone probabilistic solver for explicit 2D–3D noise models. Moreover, at run-time, AMIS requires updating all historical particle weights against an evolving proposal. Thus, computational cost grows quadratically with sampling iterations and may not be well suited for our use case.


\textbf{Uncertainty evaluation in PnP.} Despite growing interest, standardized protocols for evaluating PnP uncertainty remain underdeveloped. Existing benchmarks primarily assess pose accuracy \cite{ferraz2014leveraging, mlpnp, vakhitov2021uncertainty}, evaluating uncertainty only through qualitative visual inspection \cite{chen2022epro, richter2019towards}. Their focus on nominal settings further limits assessment of uncertainty reliability in failure cases, such as pose ambiguities or ill-conditioned optimization.

\textbf{Applications of Stein methods in robotics.} SVI \cite{liu2016stein} provides a non-parametric method to approximate complex distributions by deterministically transporting particles. While SVI has been applied in robotics from perception \cite{maken2021stein}, planning \cite{lee2025stamp}, to control \cite{pmlr-v155-lambert21a}, its application to visual PnP solvers remains unexplored, SteinPnP bridges this gap by extending SVN to the PnP problem for pose uncertainties.

\section{Background}
\label{sec:background}

\subsection{The Perspective-n-Point Problem}
\label{sec:pnp_problem}

Given 3D points $\{X_m\}_{m=1}^N$ and corresponding image coordinates $\{u_m\}_{m=1}^N$, the PnP problem estimates the camera pose parameters $\theta$. A pose defines a rigid transformation \((R(\theta), t(\theta)) \in \mathrm{SE}(3)\) from world to camera coordinates:
\begin{equation} 
X_m^c(\theta)=R(\theta)X_m+t(\theta). 
\end{equation}
Assuming calibrated image coordinates, the reprojection residual of a correspondence $(u_m,X_m)$ is 
\begin{equation} 
r_m(\theta)=u_m-\pi\!\left(X_m^c(\theta)\right), \end{equation} 
Where $\pi(\cdot)$ denotes the perspective projection operator in calibrated image coordinates. The PnP estimate is obtained by minimizing the reprojection residuals over all correspondences.

\subsection{PnP under Input Uncertainty}
\label{sec:uncertain_pnp}

For uncertainty estimation, we consider uncertainty in both the 2D image observations and the 3D points. Following previous work \cite{vakhitov2021uncertainty}, both are assumed to be corrupted by zero-mean additive Gaussian noise, i.e., \(X_m = \bar X_m + \xi_m\) and \(u_m = \bar u_m + \zeta_m\), where \(\bar X_m\) and \(\bar u_m\) are the unknown true positions. The noise terms $\zeta_m \sim \mathcal{N}(0, \Sigma_{u_m})$ and $\xi_m \sim \mathcal{N}(0, \Sigma_{X_m})$ are mutually independent. Then, for $N$ correspondences, a PnP solver now minimizes an uncertainty-aware objective:
\begin{equation}
  \mathcal{L}(\theta) = \frac{1}{2} \sum_{m=1}^{N}
    r_m(\theta)^{\top}\, \Sigma_{r_m}(\theta)^{-1}\, r_m(\theta).
\end{equation}
To the residual term $r_m(\theta)$, stochastic elements are added:
\begin{equation}
  \label{eq:noisy_residual}
  r_m(\theta) =
    \underbrace{\bar u_m - \pi\big(R\,\bar X_m + t\big)}_{\text{deterministic}}
    + \underbrace{\zeta_m - J_{\pi}\, R\, \xi_m}_{\text{stochastic}}
    + \mathcal{O}\big(\lVert \xi_m \rVert^{2}\big),
\end{equation}
where \(J_{\pi}\) is the Jacobian of \(\pi\). The residual term at the ground truth \(\theta^{\mathrm{gt}}\) follows a zero mean gaussian $r_m(\theta^{\mathrm{gt}}) \sim \mathcal{N}\big(0, \Sigma_{r_m}(\theta^{\mathrm{gt}})\big)$ where:
\begin{equation}
      \Sigma_{r_m}(\theta) = \Sigma_{u_m}
    + J_{\pi}\, R\, \Sigma_{X_m}\, R^{\top} J_{\pi}^{\top}.
\end{equation}
We refer readers to \cite{vakhitov2021uncertainty} for the detailed derivation. The uncertainty-aware loss term intuitively down-weights the correspondences whose
3D point or image measurement is more noisy. This formulation is used within SteinPnP.

\subsection{Stein Variational Inference}

Stein variational inference is an approximate Bayesian inference method that represents a target distribution with a set of particles, providing a non-parametric approximation. This method iteratively transports a set of $P$ particles \(\Theta^{0} = \{\theta_i^{0}\}_{i=1}^{P}\) drawn from an initial density \(q^{0}\). At iteration \(\ell\) each particle is pushed through the transport map
\(\mathcal{T}_{\ell}(\theta) = \theta + \epsilon\phi^{\ell}(\theta)\) which is a perturbation of the identity by a smooth vector field \(\phi^{\ell}\) with step size \(\epsilon\), giving the update:
\begin{equation}
  \label{eq:svgd_update}
  \theta_i^{\ell+1} \leftarrow \theta_i^{\ell}
    + \epsilon\, \phi^{\ell}(\theta_i^{\ell}), \quad i = 1,\dots,P .
\end{equation}
The vector field \(\phi^{\ell}\) defines the steepest-descent direction for minimizing the KL divergence between the particle approximation and the target distribution. Through iterative updates, the particles converge to the target distribution.

The field is typically chosen in a closed form as:
\begin{equation}
  \label{eq:phi_star}
  \phi^{\ell\star}(\theta) = \mathbb{E}_{\theta' \sim q^{\ell}}
    \left[ k(\theta',\theta)\, s_p(\theta')
    + \nabla_{\theta'} k(\theta',\theta) \right].
\end{equation}
The score function $s_p(\theta) = \nabla_\theta \log p(\theta)$ is the gradient of the log density whereas a kernel function \(k(\theta,\theta')\) measures the similarity between particles. The expectation term can be approximated empirically with the particles. Intuitively, \eqref{eq:phi_star} combines two forces: an \textit{attraction term} driven by the score \(s_p\), pulling particles towards high-probability regions, and a \textit{repulsion term} driven by the kernel gradient, spreading particles apart to maintain diversity and prevent local collapse. 

Following the terminology in optimization, Stein's method based on first-order gradients is known as Stein Variational Gradient Descent (SVGD) \cite{liu2016stein}. Extending the method to incorporate second-order information yields Stein Variational Newton (SVN) \cite{detommaso2018stein}, which exploits Hessian information $G(\theta)$ to better capture the local geometry of the target distribution. As a result, SVN typically converges more rapidly than SVGD, although it incurs additional computational cost. Readers are referred to the original works on SVGD and SVN for further theoretical details and convergence analyses. Importantly, applying SVN to the PnP problem therefore requires deriving appropriate expressions for the kernel, score function, and Hessian, which we present next.

\captionsetup{font=small, skip=4pt}\section{SteinPnP: Mathematical Formulation}
\label{sec:stein_pnp}

To obtain a distribution over the camera pose, we
cast PnP as a Bayesian inference problem
(Sec.~\ref{sec:pnp_as_bayesian_inference}) and approximate its intractable
posterior by a set of particles evolved with SVN (Sec.~\ref{sec:posterior_approx}). We also show how to recover a MAP pose from this non-parametric representation (Sec.~\ref{sec:map}).

\subsection{PnP as a Bayesian Inference Problem}
\label{sec:pnp_as_bayesian_inference}

We parametrise the camera pose by \(\theta=\left[ \theta_{1:3} \ \theta_{4:6} \right]\), where
\(\theta_{1:3}\) is the translation and \(\theta_{4:6}\) the orientation in
Euler angles. Given a set of 2D--3D correspondences
\(\mathcal{D} = \{(u_m, X_m)\}_{m=1}^{N}\), we seek the posterior distribution:
\begin{equation}
  \label{eq:generic_posterior}
  p(\theta \mid \mathcal{D})
    = \frac{p(\mathcal{D} \mid \theta)\, p(\theta)}{Z},
\end{equation}
where \(p(\theta)\) is the prior, \(p(\mathcal{D}\mid\theta)\) the likelihood
of observing the correspondences under a pose \(\theta\), and
\(Z = \int p(\mathcal{D} \mid \theta)\, p(\theta)\, \mathrm{d}\theta\) the
normalization constant, which is typically intractable. Assuming mutual independence of the residual terms, the likelihood follows from the probabilistic description of the residual \eqref{eq:noisy_residual}. We use a uniform prior \(p(\theta)\). Then,
\begin{equation}
    \log p(\theta \mid \mathcal{D}) = -\frac{1}{2} \sum_{m=1}^N \left( r_m^\top \Sigma_{r_m}^{-1} r_m + \log \det \Sigma_{r_m} \right) + \text{const},
      \label{eq:log_posterior}
\end{equation}
where $r_m$ and $\Sigma_{r_m}$ implicitly depend on $\theta$. Crucially, the normalizer $Z$ is $\theta$-independent and vanishes under the score function $s_p(\theta) = \nabla_\theta \log p(\theta \mid \mathcal{D})$, along with other terms.

\subsection{Approximation of the Posterior Distribution with SVN}
\label{sec:posterior_approx}

Having defined the target distribution, we approximate it by a set of \(P\)
particles \(\Theta^{\ell} = \{\theta_i^{\ell}\}_{i=1}^{P}\), each representing a
pose in the parametrisation above. Starting from an initial ensemble
\(\Theta^{0}\), we iteratively advance the particles:
\begin{equation}
  \label{eq:stein_pnp_update}
  \theta_i^{\ell+1} \leftarrow \theta_i^{\ell}
    + \epsilon\, \phi^{\ell\mathrm{N}}(\theta_i^{\ell}),
    \quad i = 1,\dots,P .
\end{equation}
Again, the Newton direction \(\phi^{\ell\mathrm{N}}\) requires three ingredients: the score
\(s_p(\theta)\), the Hessian \(G(\theta)\), and a kernel
\(k(\theta, \theta')\).

Throughout, \(\partial_a := \partial / \partial \theta_a\) denotes
differentiation with respect to the \(a\)-th pose coordinate,
\(a \in \{1, \dots, d\}\) with \(d = 6\). Accordingly,
\(J_{r_m} = \partial r_m / \partial \theta \in \mathbb{R}^{2 \times 6}\) is the
residual Jacobian, whose \(a\)-th column is \(\partial_a r_m\), and
\(\partial_a \Sigma_{r_m} \in \mathbb{R}^{2 \times 2}\) is the entrywise
derivative of the residual covariance.

\paragraph{Score} The score function is defined as the gradient of the log-posterior. Since \(\theta\) enters both the residual and its covariance, it splits into two contributions:
\begin{align}
  \label{eq:score}
  \big[ s_p(\theta) \big]_a &=
    -\underbrace{\sum_{m=1}^{N}
      (\partial_a r_m)^{\top} \Sigma_{r_m}^{-1} r_m}
      _{\textbf{(R)}\ \text{residual path}}\\
      &\quad- \underbrace{\frac{1}{2} \sum_{m=1}^{N} \operatorname{tr}\!\Big[
      \big( \Sigma_{r_m}^{-1}
      - \Sigma_{r_m}^{-1} r_m r_m^{\top} \Sigma_{r_m}^{-1} \big)\,
      \partial_a \Sigma_{r_m} \Big]}
      _{\textbf{(C)}\ \text{covariance path}} \nonumber.
\end{align}
Term \textbf{(R)} is the weighted reprojection gradient, whereas term \textbf{(C)} models covariance variations and is often discarded when \(\Sigma_{r_m}\) is held fixed in certain optimization schemes.

\paragraph{Hessian} SVN requires the Hessian to be positive semi-definite. As a surrogate for the Hessian, we use the Fisher information matrix $G_{\text{Fisher}}$, \emph{i.e.}, the expected outer product of the score: \(G(\theta) = \mathbb{E}_{\mathcal{D} \sim p(\cdot \mid \theta)} \big[ s_p(\theta)\, s_p(\theta)^{\top} \big]\). The Fisher also captures the expected curvature of the log-posterior at a given pose without requiring second derivatives of the projection. In our formulation, the Fisher admits the closed form:
\begin{align}
    \label{eq:fisher}
  G_{\text{Fisher}_{ab}}(\theta) &=
    \underbrace{\sum_{m=1}^{N}
      \big[ J_{r_m}^{\top} \Sigma_{r_m}^{-1} J_{r_m} \big]_{ab}}
      _{\textbf{(R)}}\\
    &\quad + \underbrace{\frac{1}{2} \sum_{m=1}^{N} \operatorname{tr}\!\big(
      \Sigma_{r_m}^{-1} \partial_a \Sigma_{r_m}\,
      \Sigma_{r_m}^{-1} \partial_b \Sigma_{r_m} \big)}
      _{\textbf{(C)}},\nonumber
\end{align}
where the residual path yields the Gauss--Newton matrix, and the covariance path includes an additional curvature term.

\paragraph{Kernel} We use a Gaussian curvature-scaled (RBF) kernel ~\cite{detommaso2018stein}:
\begin{equation}
  \label{eq:scaled_kernel}
  k^{\ell}(\theta, \theta') = \exp\!\left( -\tfrac{1}{d}\,
    \lVert \theta - \theta' \rVert_{\bar G^{\ell}}^{2} \right),
  \qquad
  \lVert v \rVert_{\bar G^{\ell}}^{2} = v^{\top} \bar G^{\ell}\, v ,
\end{equation}
which uses the ensemble-averaged Fisher information \(\bar G^{\ell} = \tfrac{1}{P} \sum_{i=1}^{P} G_{\text{Fisher}}(\theta_i^{\ell})\) as its metric. It comes at no additional cost, since each \(G_{\text{Fisher}}(\theta_i^{\ell})\) is already formed for the Newton step, and it lets a single kernel act consistently on translation and rotation despite their differing units. The kernel is broader along low-curvature directions and narrower along high-curvature directions, so that attraction and repulsion act uniformly over a strongly anisotropic target. The scaling \(1/d\) replaces tuning of kernel scales.

\subsection{Maximum a Posteriori Estimation}
\label{sec:map}

After enough iterations $T$, the converged particles $\Theta^{T}$ approximate
the shape of the posterior and therefore occupy the regions of high density.
We leverage this to obtain a robust estimate of the \emph{maximum a
posteriori} (MAP) pose $\hat\theta^{\star}$, i.e., the pose with the highest
probability of generating the data $\mathcal{D}$. Starting from each
converged particle \(\theta_i^{T}\) of the particle set \(\Theta^T\), we run a few iterations of a damped Newton
(Levenberg--Marquardt-style) ascent on the log posterior
\eqref{eq:log_posterior}, using the score $s_p(\theta)$ and the curvature
$G(\theta)$ already computed for the Newton step and obtain an ascended set of particles condensed tightly around the modes \(\hat \Theta = \big\{\hat \theta_i\big\}_{i=1}^P\), and keep the particle reaching the highest probability as the MAP estimate \(\hat \theta^\ast = \arg \max_{\theta \in \hat \Theta} \{\log p(\theta \,|\, \mathcal{D})\}\). Due to the multiple starting points, this gives a high probability of
finding the global maximum at little extra cost.


\captionsetup{font=small, skip=4pt}\section{SteinPnP in Practice}
\label{sec:stein_pnp_practice}

\begin{figure}[t]
  \centering
  \hspace*{-1cm}
  \includegraphics[width=0.6\linewidth]{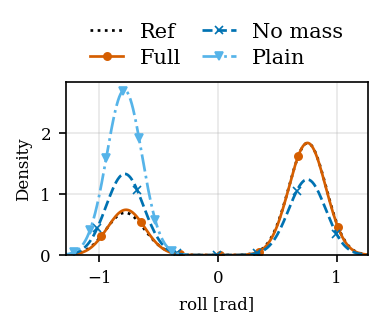}
  \caption{\footnotesize Marginal roll KDE posterior for a planar PnP instance.}
  \label{fig:pitfalls}
  \vspace*{-5mm}
\end{figure}

In practice, however, SVN can struggle with the multimodal posteriors typical of planar or ambiguous PnP configurations. When modes are separated by wide, low-probability valleys, standard SVN suffers from two distinct issues:
\begin{enumerate}
    \item \textbf{Mode Collapse:} Particles initialized near a single mode fail to traverse low-density regions to discover secondary modes (\emph{Plain} curve in Fig.~\ref{fig:pitfalls}). Therefore, some of the modes are not discovered by the particles.
    \item \textbf{Incorrect Mass:} Mass allocation among modes is inherited from initialization rather than the target distribution (\emph{No mass} curve in Fig.~\ref{fig:pitfalls}). Thus, the modes are covered but the ratio of particles between them is biased.
\end{enumerate}
To address these challenges, we propose a practical SVN algorithm for the PnP problem. Specifically, we propose a randomized P3P initialization (Sec.~\ref{sec:p3p_init}) to promote mode coverage, alongside an explicit mode mass refinement step (Sec.~\ref{sec:mass_refinement}) to reallocate particles according to mode probabilities. Together, these modifications enable SteinPnP (\emph{Full} curve) to better capture the true posterior, thereby providing more faithful quantification of the pose uncertainty.

\subsection{Mode-Covering Initialization via Randomized P3P}
\label{sec:p3p_init}
To ensure proper mode coverage, we require the initial ensemble $\Theta^0$ of $P$ particles to place a portion of its population in each mode. We leverage P3P as a computationally cheap way to probe the different modes, through a Monte Carlo--like procedure: for each particle $\theta_i^0$, we draw a unique\footnote{The number of distinct 3-subsets is $\binom{N}{3} = \frac{N!}{3!(N-3)!}$, which for small $N$ can be smaller than $P$. When it is, we resample already-drawn subsets, relying on the noise below to keep the resulting particles diverse.} subset $\mathcal{S}_i \subset \mathcal{D}$, $|\mathcal{S}_i| = 3$, and corrupt each of its 2D and 3D points with noise drawn from their respective covariances, $\tilde{u}_m \sim \mathcal{N}(u_m, \Sigma_{u_m})$ and $\tilde{X}_m \sim \mathcal{N}(X_m, \Sigma_{X_m})$ for $(u_m, X_m) \in \mathcal{S}_i$. $\theta_i^0$ is then set to whichever of the (up to four) P3P solutions on $\{(\tilde{u}_m, \tilde{X}_m)\}$ has the smallest mean squared reprojection error over $\mathcal{D}$.

\subsection{Mode Mass Refinement}
\label{sec:mass_refinement}
To improve particle mass allocation across modes, our mode-mass refinement step (a) identifies the modes and assigns each particle to its mode; if multiple modes are detected, it then (b) estimates their probability masses via importance sampling and (c) resamples the particles accordingly before performing additional SVN iterations.

\subsubsection{Mode detection via post-ascent clustering}
To detect modes, we first move each particle to its local log-posterior maximum using a brief gradient ascent (the ascended particles $\hat{\Theta}$ of Sec.~\ref{sec:map} can be reused). Particles belonging to the same mode contract into dense clusters separated by wide gaps, such that every intra-mode distance is shorter than any inter-mode distance. We then connect the ascended particles using a minimum spanning tree (MST), \emph{i.e.}, a tree connecting all $P$ particles through $P-1$ edges of minimum total weight, where the weight $w_i$ of edge $i$ is the geodesic rotation distance between the two particles it connects. Because the MST minimizes the total edge weight, it connects the $C$ modes through only $C-1$ ``bridge'' edges, the minimum number required to keep the tree connected, and removing these bridges recovers the modes.

Since the bridges are few ($C-1 < (P-1)/2$ because $P \gg C$), the median edge weight $\tilde{w} = \operatorname{med}_i w_i$ is an intra-mode edge and characterizes the within-mode scale. Following a global variant of Zahn's inconsistent-edge criterion~\cite{zahn1971gestalt}, we remove all edges satisfying $w_i > 10 \,\tilde{w}$ and take the resulting connected components as the modes $\{\mathcal{C}_c\}_{c=1}^{C}$, each centered at its highest-posterior particle $m_c$. This recovers the exact partition provided that intra-mode edges are shorter than $10 \,\tilde{w}$ and modes are separated by more than $10 \,\tilde{w}$. The factor is intentionally kept low, because fused modes cannot subsequently be separated, whereas split modes are merged back whenever the log-posterior along the geodesic connecting their centers falls by less than 5~nats below the lower peak.

\subsubsection{Mode mass approximation}
For each mode $c$, we estimate its share of the posterior mass by importance sampling against a proposal distribution $q_c$. We approximate the geometry of mode $c$ with the within-mode-averaged Fisher information, evaluated on the pre-ascent particles $\theta_i^{T}$ assigned to $\mathcal{C}_c$:
\begin{equation}
  q_c(\theta) = \mathcal{N}\big(\theta \mid m_c,\, \bar{G}_c^{-1}\big),
  \qquad
  \bar{G}_c = \frac{1}{|\mathcal{C}_c|} \sum_{i \in \mathcal{C}_c} G(\theta_i^{T}).
\end{equation}
Writing $\gamma(\theta) = p(\mathcal{D} \, | \, \theta)$, the unnormalized posterior reduces to the likelihood since the prior is uniform. The mass contained in mode $c$ is estimated as
\begin{gather}
  Z_c = \int_{\Omega_c} \frac{\gamma(\theta)}{q_c(\theta)}\,q_c(\theta)\, d\theta
      = \mathbb{E}_{\theta \sim q_c}\!\left[\frac{\gamma(\theta)}{q_c(\theta)}\right]
\end{gather}
where $\Omega_c$ denotes the region associated with mode $c$. For stability reasons, we compute \(\log Z_c\) instead of \(Z_c\).

\subsubsection{Resampling and refinement}
The estimated masses per mode $Z_c$ set the target population of each mode $P_c$, \emph{i.e.},
\begin{equation*}
  P_c = \big\lfloor P\, \pi_c \big\rfloor,
  \qquad
  \pi_c = \frac{Z_c}{\sum_{c'} Z_{c'}}
        = \operatorname*{softmax}_c\big(\log Z_c\big).
\end{equation*}
The softmax is used for normalizing the log-masses  \(\log Z_c\) while ensuring numerical stability. $P - \sum_c P_c$ leftover particles are assigned by largest remainder, so that $\sum_c P_c = P$ exactly. The correction is applied to the SVN ensemble $\Theta^{T}$, each particle inheriting the mode of its ascended image $\hat\theta_i$, and each mode is corrected by exactly the amount it is off target rather than rebuilt. An over-represented mode ($|\mathcal{C}_c| > P_c$) has $|\mathcal{C}_c| - P_c$ of its particles removed, chosen uniformly at random; an under-represented one ($|\mathcal{C}_c| < P_c$) keeps every particle it already has and is topped up with $P_c - |\mathcal{C}_c|$ draws from the proposal $q_c = \mathcal{N}\big(m_c,\, \bar G_c^{-1}\big)$ already built for the mass estimate. The particles SVN has already fitted to their mode are thus preserved and only the newly drawn ones need to relax, so that a handful of further SVN iterations (Sec.~\ref{sec:staged_updates}) suffices to re-equilibrate the ensemble.

\subsection{Staged Updates for Efficiency}
\label{sec:staged_updates}

Computing the score \eqref{eq:score} and Fisher information matrix \eqref{eq:fisher} is costly, mainly because the covariance path \textbf{(C)} requires evaluating \(\partial_a\Sigma_{r_m}\) for every residual and pose dimension. We therefore perform most iterations using a surrogate update that treats \(\Sigma_{r_m}\) as pose-independent while refreshing it between iterations. This variational-inference analogue of iteratively reweighted least squares \cite{vakhitov2021uncertainty} retains only the residual path \textbf{(R)}. These inexpensive updates bring the ensemble close to the target posterior, after which \(10\) full iterations with \textbf{(C)} restored eliminate the remaining discrepancy. The resulting schedule thus matches the fully derived update at a fraction of its cost.


\subsection{Robustness to a Miscalibrated Noise Model}
\label{sec:miscalibration}

To recap, we assumed a noise model for 3D points and 2D observations. The two paths of SteinPnP, namely the residual path and the covariance path, behave differently when the noise model is miscalibrated. Suppose the residuals remain zero-mean at the ground-truth pose but the scale of the covariance is misspecified by a common factor $\alpha$, i.e.,
\begin{equation}
  \label{eq:miscalibration}
  \mathbb{E}_{\mathcal{D}}\big[r_m\big] = 0,
  \qquad
  \mathbb{E}_{\mathcal{D}}\big[r_m r_m^{\top}\big] = \alpha\, \Sigma_{r_m},
  \qquad \alpha \neq 1 .
\end{equation}
Let $s_R$ and $s_C$ denote the residual and covariance paths of the score \eqref{eq:score}. Taking the expectation at the ground truth $\theta^{\mathrm{gt}}$:
\begin{align}
  \mathbb{E}_{\mathcal{D}}\big[s_R(\theta^{\mathrm{gt}})\big]_a
    &= -\sum_{m=1}^{N} (\partial_a r_m)^{\top} \Sigma_{r_m}^{-1}\,
       \mathbb{E}_{\mathcal{D}}\big[r_m\big] = 0,
       \label{eq:bias_R} \\
  \mathbb{E}_{\mathcal{D}}\big[s_C(\theta^{\mathrm{gt}})\big]_a
    &= -\frac{1}{2} \sum_{m=1}^{N} \operatorname{tr}\!\Big[
       \big( \Sigma_{r_m}^{-1} - \alpha\, \Sigma_{r_m}^{-1} \big)\,
       \partial_a \Sigma_{r_m} \Big] \notag \\
    &= \frac{\alpha - 1}{2} \sum_{m=1}^{N}
       \partial_a \log \det \Sigma_{r_m} \;\neq\; 0 .
       \label{eq:bias_C}
\end{align}
The residual path is therefore unbiased: whatever the scale of the covariance matrices, the ground truth remains a stationary point in expectation. The covariance path is not. 

On the other hand, the direction of the displacement is interpretable: for \(\alpha > 1\), when the residuals are larger than the model admits, the bias points along \(+\nabla_\theta \sum_m \log \det \Sigma_{r_m}\), that is, towards poses whose predicted residual covariances have larger volume; the pose thus absorbs an error induced by the noise model. Concretely, it favours reducing the depth at which the 3D covariances are projected, or rotating the camera so that their elongated axes become fronto-parallel and project to wider image ellipses; $\alpha < 1$ biases the opposite way. This motivates us to run the \textit{MAP} iterations (Sec.~\ref{sec:map}) using the residual path only, which empirically resulted in higher accuracy. In contrast, in the refinement iterations of Sec.~\ref{sec:staged_updates}, we also use the covariance path: even though the bias of \eqref{eq:bias_C} corrupts the location of the modes, $s_C$ helps to capture the complete curvature of the log-posterior. 
\section{Experiments}
\label{sec:experiments}

We claim the following: First, coplanar point configurations can induce multimodal PnP posteriors that traditional Gaussian PnP solvers cannot represent. Second, SteinPnP can capture multimodal posteriors in coplanar settings, while converging to a unimodal distribution in non-planar scenes. Third, SteinPnP achieves these results at a fraction of MCMC's computational cost without compromising accuracy. We validate these claims on synthetic and real-world datasets, including TUM RGB-D and an AprilTag sequence.


\subsection{Setup: Baselines, Protocols, Implementation $\&$ Metrics}
\label{sec:exp:setup}

We evaluate SteinPnP against state-of-the-art uncertainty-aware PnP solvers: MLPnP \cite{mlpnp}, EPnPU \cite{vakhitov2021uncertainty}, and DLSU \cite{vakhitov2021uncertainty}. Both CEPnP and MLPnP model only 2D uncertainty; we retain MLPnP for its native pose covariance output.
 While all baselines leverage input uncertainty to improve pose accuracy, only MLPnP natively outputs pose uncertainty. To extend the availability of baselines and ensure a fair comparison, we extend EPnPU and DLSU by propagating input uncertainty using MLPnP's linear propagation strategy. For initialization, OpenCV's PnP-RANSAC provides the initial pose for EPnPU, DLSU, and SteinPnP alike. To isolate the impact of our technical contributions, we also conduct ablation studies comparing our full \textit{SteinPnP} pipeline against variants without mode mass refinement (\textit{no mass}; Sec.~\ref{sec:mass_refinement}), without covariance path refinement iterations (\textit{no ref}; Sec.~\ref{sec:staged_updates}), and a \textit{Vanilla} baseline applying standard SVGD directly to the PnP problem. Its MAP estimate is simply taken as the particle with the highest log-posterior \eqref{eq:log_posterior}  (Sec.~\ref{sec:stein_pnp}).


Evaluating uncertainty is challenging due to the lack of ground truth. A standard practice in Bayesian inference is to evaluate uncertainty estimates against a reference posterior drawn from sampling algorithms like MCMC (\textit{EMCEE}, \cite{Foreman_Mackey_2013}) or dynamic nested sampling (\textit{Dynesty} \cite{speagle2020dynesty}). Although these sampling methods accurately capture true posteriors and relative mode weights, their run-time is too slow for applications in robotics. Thus, these gold-standard sampling methods are used offline as ground-truth references that their estimates aim to match. Our implementation details are as follows. For multimodal coplanar scenes, we generate reference posteriors using \textit{Dynesty} with $2000$ samples. For unimodal non-planar scenes, we employ a faster \texttt{emcee} ensemble sampler ($32$ walkers, $300$ burn-in, and $2000$ production steps). Evaluating these reference samplers against held-out samples establishes an empirical performance floor for all methods, extending standard posterior validation practices \cite{maken2021stein, wilson2022evaluating} from the literature to the PnP setting. 


SteinPnP uses (100) particles and performs (140) iteratively reweighted updates (Sec.~\ref{sec:staged_updates}) with step size (2.2), (10) \textit{MAP} iterations (Sec.~\ref{sec:map}), mode-mass refinement (Sec.~\ref{sec:mass_refinement}), and (10) full-gradient updates (Sec.~\ref{sec:staged_updates}) with step size (1.0). The ablation baseline, \textit{SteinPnP-Vanilla}, instead runs standard SVGD for (250) full-gradient iterations with step size (0.02), without mode-covering initialization or mass refinement. All reference samplers (\textit{Dynesty} and \texttt{EMCEE}) are initialised using ground-truth-informed wide priors/initializations that cover all modes.


We report three families of quantities. \emph{Accuracy}: the rotation error $e_R$ [deg] and translation error $e_t$ [m] of each method's point estimate following \cite{vakhitov2021uncertainty}. \emph{Distribution quality}, measured against the reference sample in two ways: the unbiased $\mathrm{MMD}^2_u$, a non-parametric statistic sensitive to the full shape of the posterior, and the squared error on the posterior mean and second moment. \emph{Cost}: CPU runtime [s]. Gaussian baselines are sampled with the same number of particles as SteinPnP, so that same sample sizes are used for all methods.


\begin{figure}
    \centering
    \includegraphics[width=0.95\columnwidth]{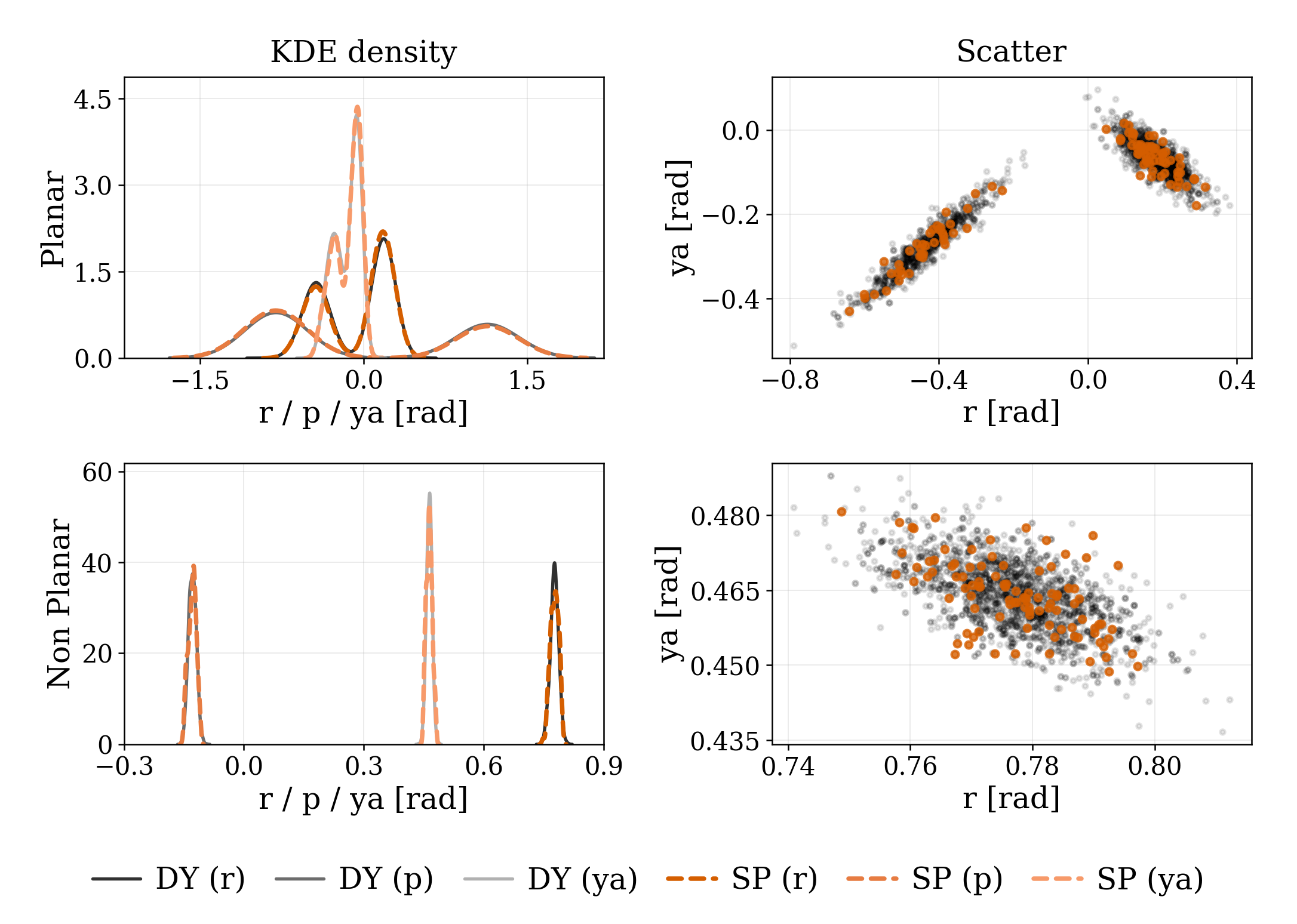}
    \caption{\footnotesize Euler angle marginals (left) and roll--yaw scatter plots (right) comparing SteinPnP against the Dynesty (reference) for coplanar (top) and non-planar (bottom) scenarios. SteinPnP accordingly captures both the bimodal and unimodal posterior.}
    \label{fig:qualitative}
    \vspace*{-5mm}
\end{figure}

\subsection{Experiments on Synthetic Data}
\label{sec:exp:synthetic}

To evaluate pose uncertainty under controlled geometries and noise models, we generate 100 PnP trials per configuration using a calibrated camera. Following Vakhitov et al.~\cite{vakhitov2021uncertainty}, non-planar points are sampled in 3D space, while planar targets follow Schweighofer and Pinz~\cite{Schweighofer2006RobustPE} using ten uniformly sampled points whose projections fall within a bounding box. We vary, separately, the depth $t_z\in\{10,15,20,25,30\}$\,m, the angle between the camera-frame plane normal and $\mathbf{t}$ in $\{0,15,30,45,60,75,90\}^\circ$, and the noise level $\sigma_{\max}\in\{1,3,5,7,9,11\}$\,px. Noise models are also varied to simulate various conditions. We use zero-mean anisotropic Gaussian noise on the 3D points and their 2D projections \cite{vakhitov2021uncertainty}. Noise ellipsoids are randomized. Given a maximum standard deviation $s$, remaining axes are sampled uniformly from $(0,s)$. Geometric sweeps assign points across ten noise levels ($1\text{--}10$\,px in 2D, $0.01\text{--}0.10$\,m in 3D), while noise sweeps sample $s$ uniformly up to $\sigma_{\max}$\,px and $0.01\,\sigma_{\max}$\,m.

\begin{figure}
    \centering
    \includegraphics[width=0.95\columnwidth]{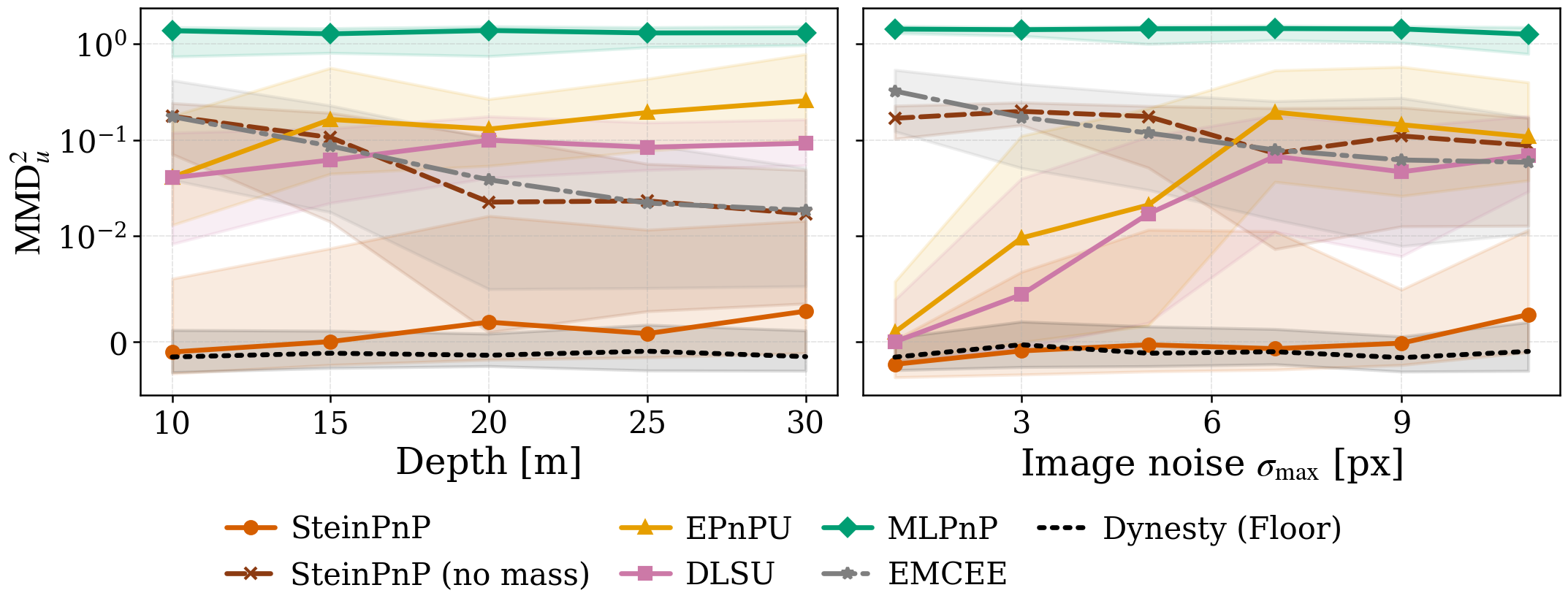}
    \caption{\footnotesize $\mathrm{MMD}^2_u$ across varying scene depth (left) and image noise $\sigma_{\max}$ (right) under planar geometry (shaded bands: IQR over $100$ trials). SteinPnP consistently tracks the reference, whereas Gaussian baselines diverge as geometric ambiguities strengthen.}
    \label{fig:sweep}
    \vspace*{-5mm}
\end{figure}

Qualitative results are depicted in Fig.~\ref{fig:qualitative} and \ref{fig:sweep}. Fig.~\ref{fig:qualitative} shows that on a non-planar scene the posterior is unimodal and tightly concentrated, whereas on a planar scene the posterior is bimodal, with one mode at the true pose and one at the ambiguous pose \cite{Schweighofer2006RobustPE}. SteinPnP produces reliable uncertainty estimates in both cases. Fig.~\ref{fig:sweep} shows how the ambiguity in coplanar scene increases with noise: At $\sigma_{\max}=1\,\mathrm{px}$, all Gaussian methods except MLPnP are close to the Dynesty reference; they then deviate monotonically as noise increases and the coplanar ambiguity strengthens. Conversely, EMCEE and SteinPnP (no mass), which tend to overrepresent the ambiguous mode without explicit mode-mass control, approach Dynesty as ambiguity strengthens, whereas SteinPnP remains near the floor throughout.


Tab.~\ref{tab:synthetic} shows quantitative results. We extend our evaluation over $1800$ coplanar and $600$ non-planar scenarios. Three readings matter. In the coplanar setting, SteinPnP is the only method whose $\mathrm{MMD}^2_u$ and moment errors sit near the Dynesty (floor). The errors of EPnPU and DLSU are one to two orders of magnitude larger. In the non-planar setting all methods except MLPnP and SteinVanilla converge on $e_R$ and $e_t$. This is expected as 
Gaussian assumption holds well in the non-planar settings. Run-time of SteinPnP is slower than Gaussian-based solvers, but an order of magnitude faster than sampling methods. 


\begin{table*}
\centering
\footnotesize
\setlength{\tabcolsep}{4pt}
\caption{\footnotesize Quantitative results from synthetic benchmark. We report median (IQR) across $1800$ coplanar and $600$ non-planar scenarios. Dynesty is the reference. SteinPnP uniquely matches Dynesty on distributional fidelity ($\mathrm{MMD}^2_u$) across both settings, while maintaining competitive accuracy. SteinPnP is slower than Gaussian-based solvers but orders of magnitude fastest among non-parametric methods. Freq Global Max denotes the frequency of identifying the global posterior maximum.}
\label{tab:synthetic}
\resizebox{\textwidth}{!}{%
\begin{tabular}{lrrrrrrr}
\toprule
Method & Freq Global Max & $e_R$ [deg] & $e_t$ [m] & $\mathrm{MMD}^2_u$ & Mean error & Second-moment error & Time [s] \\
\midrule
\multicolumn{8}{l}{\textbf{Planar}} \\
SteinPnP & {\boldmath $1\;(0.0103)$} & {\boldmath $8.17\;(64.1)$} & {\boldmath $0.462\;(0.894)$} & {\boldmath $-8.95\mathrm{e}{-5}\;(0.0105)$} & {\boldmath $0.0143\;(0.0822)$} & {\boldmath $15.4\;(113)$} & $0.982\;(0.144)$ \\
SteinPnP (no mass) & {\boldmath $1\;(0.0103)$} & $8.17\;(64.1)$ & $0.462\;(0.902)$ & $0.0562\;(0.184)$ & $0.344\;(1.04)$ & $235\;(744)$ & $0.665\;(0.0701)$ \\
SteinPnP-Vanilla & $0.818\;(0.0546)$ & $10.4\;(75.8)$ & $0.569\;(1.07)$ & $0.216\;(0.372)$ & $0.99\;(2.53)$ & $757\;(3.16\mathrm{e}{3})$ & $1.06\;(0.114)$ \\
EPnPU & $0.675\;(0.16)$ & $8.45\;(52)$ & $0.495\;(1.1)$ & $0.116\;(0.33)$ & $0.576\;(3.52)$ & $568\;(4.01\mathrm{e}{3})$ & {\boldmath $6.5\mathrm{e}{-3}\;(5.23\mathrm{e}{-4})$} \\
DLSU & $0.883\;(0.0478)$ & $8.43\;(67.1)$ & $0.462\;(0.887)$ & $0.0652\;(0.128)$ & $0.3\;(1.35)$ & $295\;(1.68\mathrm{e}{3})$ & $0.14\;(0.0103)$ \\
MLPnP & $0.524\;(0.0541)$ & $134\;(116)$ & $19.4\;(37.7)$ & $1.36\;(0.56)$ & $386\;(1.64\mathrm{e}{3})$ & $1.1\mathrm{e}{4}\;(9.17\mathrm{e}{4})$ & $0.0115\;(8.57\mathrm{e}{-4})$ \\
EMCEE & -- & -- & -- & $0.0444\;(0.184)$ & $0.305\;(1.15)$ & $217\;(823)$ & $11.9\;(0.82)$ \\
Dynesty (floor) & -- & -- & -- & $-9.91\mathrm{e}{-4}\;(3.84\mathrm{e}{-3})$ & $5.8\mathrm{e}{-3}\;(0.0212)$ & $5.8\;(32.1)$ & $34.7\;(5.72)$ \\
\midrule
\multicolumn{8}{l}{\textbf{Non Planar}} \\
SteinPnP & {\boldmath $1\;(0)$} & {\boldmath $0.294\;(0.304)$} & {\boldmath $0.0134\;(0.0156)$} & $-2.21\mathrm{e}{-3}\;(1.48\mathrm{e}{-3})$ & $3.7\mathrm{e}{-7}\;(1.04\mathrm{e}{-6})$ & $4.57\mathrm{e}{-5}\;(1.51\mathrm{e}{-4})$ & $1.3\;(0.634)$ \\
SteinPnP (no mass) & {\boldmath $1\;(0)$} & $0.296\;(0.306)$ & $0.0139\;(0.0158)$ & {\boldmath $-3.45\mathrm{e}{-3}\;(8.74\mathrm{e}{-4})$} & $3.52\mathrm{e}{-7}\;(9.85\mathrm{e}{-7})$ & $4.37\mathrm{e}{-5}\;(1.21\mathrm{e}{-4})$ & $1.05\;(0.418)$ \\
SteinPnP-Vanilla & {\boldmath $1\;(0)$} & $0.306\;(0.31)$ & $0.0139\;(0.0156)$ & $-2.06\mathrm{e}{-3}\;(1.38\mathrm{e}{-3})$ & {\boldmath $2.34\mathrm{e}{-7}\;(6.89\mathrm{e}{-7})$} & {\boldmath $2.74\mathrm{e}{-5}\;(8.25\mathrm{e}{-5})$} & $1.95\;(0.673)$ \\
EPnPU & {\boldmath $1\;(0)$} & $0.294\;(0.308)$ & $0.0139\;(0.0158)$ & $-4.6\mathrm{e}{-4}\;(3.88\mathrm{e}{-3})$ & $2.85\mathrm{e}{-6}\;(6.94\mathrm{e}{-6})$ & $3.28\mathrm{e}{-4}\;(8.61\mathrm{e}{-4})$ & $0.0118\;(6.78\mathrm{e}{-3})$ \\
DLSU & {\boldmath $1\;(0)$} & $0.294\;(0.308)$ & $0.0139\;(0.0158)$ & $3.28\mathrm{e}{-5}\;(3.98\mathrm{e}{-3})$ & $2.9\mathrm{e}{-6}\;(8.13\mathrm{e}{-6})$ & $3.42\mathrm{e}{-4}\;(1.01\mathrm{e}{-3})$ & $0.0681\;(0.0122)$ \\
MLPnP & {\boldmath $1\;(0)$} & $1.49\;(1.45)$ & $0.0638\;(0.0695)$ & $1.51\;(0.532)$ & $4.95\mathrm{e}{-3}\;(0.0106)$ & $0.552\;(1.23)$ & {\boldmath $0.0117\;(0.0332)$} \\
EMCEE & -- & -- & -- & $3.82\mathrm{e}{-4}\;(7.4\mathrm{e}{-4})$ & $5.51\mathrm{e}{-7}\;(1.46\mathrm{e}{-6})$ & $6.83\mathrm{e}{-5}\;(1.88\mathrm{e}{-4})$ & $18.4\;(5.79)$ \\
Dynesty (floor) & -- & -- & -- & $-8\mathrm{e}{-4}\;(3.24\mathrm{e}{-3})$ & $2.4\mathrm{e}{-6}\;(5.48\mathrm{e}{-6})$ & $2.75\mathrm{e}{-4}\;(7.1\mathrm{e}{-4})$ & $67.5\;(42.1)$ \\
\bottomrule
\end{tabular}}
\vspace*{-4mm}
\end{table*}

\begin{figure}
    \centering
    \includegraphics[width=0.95\linewidth]{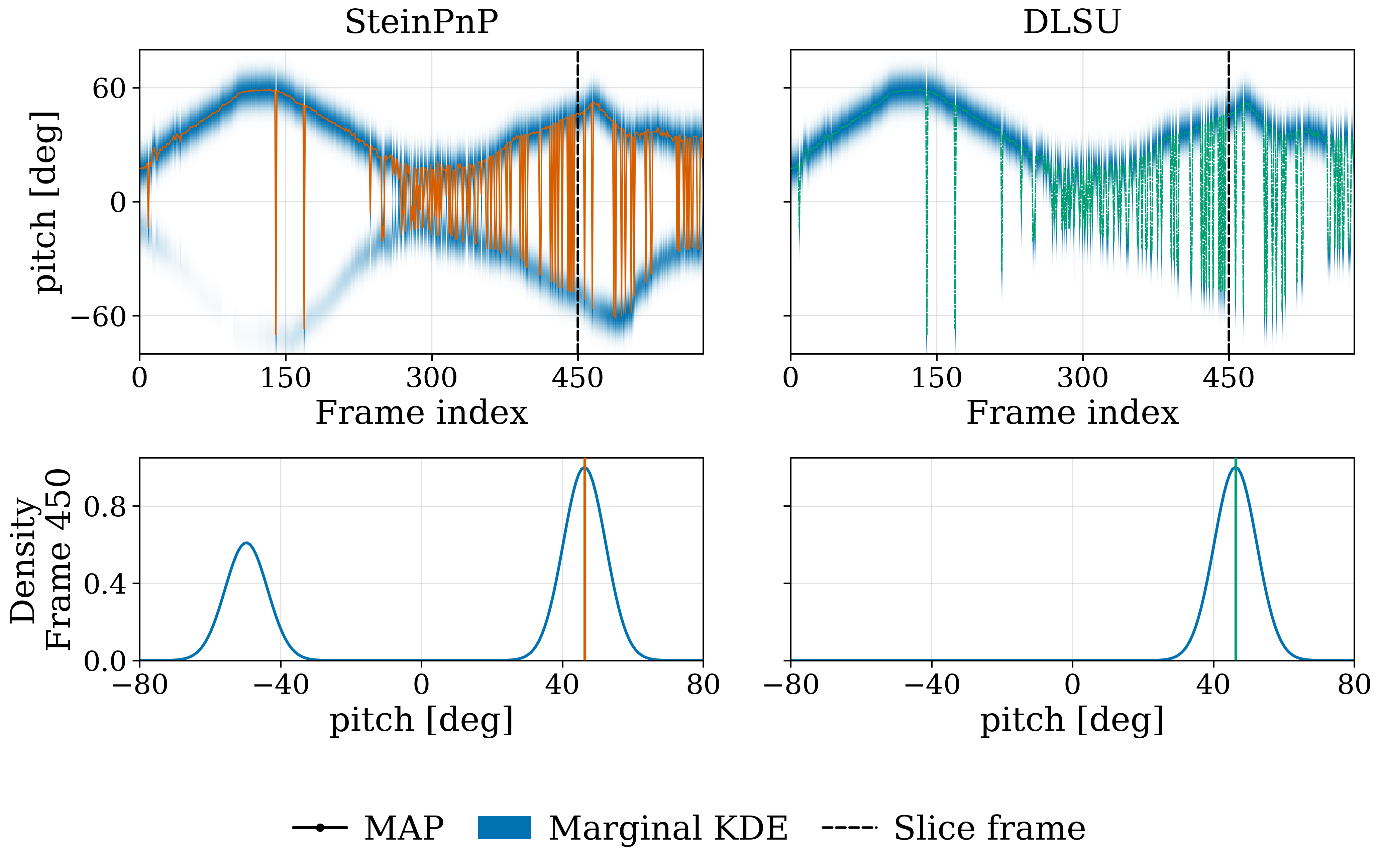}
    \caption{\footnotesize AprilTag sequence comparing SteinPnP (left) and DLSU (right). A cross-section slice at frame 450 is depicted (bottom). SteinPnP captures the bimodal distribution, whereas DLSU cannot.}
    \label{fig:apriltag}
    \vspace*{-5mm}
\end{figure}

\subsection{Experiments on Real Data}
\label{sec:exp:real}

We build PnP scenarios from real monocular image sequences with known calibration and reference trajectories. Each instance consists of 3D points reconstructed from a short temporal window of three frames and their 2D observations in a subsequent query image, and per-correspondence 2D and 3D covariances. Each reference window holds three frames with a pose distance of at least $5$\,cm between consecutive frames. The query is the next image after the window.

We extract up to $2000$ ORB features per image over an eight-level pyramid with scale factor $\kappa=1.2$. A feature detected at level $\ell$ receives the localization covariance $\Sigma^{\mathrm{dist}}_u=(\epsilon\kappa^\ell)^2 I_2$ with $\epsilon=1$\,px. Reference-frame features are filtered by a $2$-px Sampson distance computed from the known relative poses. Tracks seen in at least two reference frames are triangulated by DLT and refined by structure-only Bundle-Adjustement, from which the 3D point covariance is obtained by linear propagation of the 2D point covariance. Only correspondences with a query reprojection error below $3$\,px under the reference pose are retained. We select up to $50$ correspondences discarding instances with fewer than $10$ surviving correspondences.

Tab.~\ref{tab:real} reports results on the TUM sequences. The reconstructed scenes are non-planar. Similar to synthetic data, we observe that all methods produce similar accuracy. Interestingly, MLPnP is competitive here, which may be due to the low 3D uncertainty from close-range scenes and intra-frame pose distance over $5cm$. The distributional metrics separate them nonetheless: SteinPnP is two orders of magnitude closer to our references in $\mathrm{MMD}^2_u$ than any Gaussian baseline. We believe that, even where a Gaussian assumption holds, the linearized covariances of the baselines are noticeably wider.


\begin{table*}
\centering
\footnotesize
\setlength{\tabcolsep}{4pt}
\caption{\footnotesize Real-world evaluation on non-planar TUM RGB-D sequences reported as median (IQR). MCMC is used as the reference. Similar to before, SteinPnP achieves an order-of-magnitude lead over all baselines on distributional metrics while maintaining competitive accuracy. SteinPnP is again faster than sampling methods. Ablation studies show the contribution of each design choice within SteinPnP.}
\label{tab:real}
\resizebox{\textwidth}{!}{%
\begin{tabular}{lrrrrrr}
\toprule
Method & $e_R$ [deg] & $e_t$ [m] & $\mathrm{MMD}^2_u$ & Mean error & Second-moment error & Time [s] \\
\midrule
SteinPnP & {\boldmath $0.282\;(0.242)$} & $5.61\mathrm{e}{-3}\;(5.09\mathrm{e}{-3})$ & {\boldmath $6.83\mathrm{e}{-4}\;(4.34\mathrm{e}{-3})$} & {\boldmath $1.76\mathrm{e}{-5}\;(1.58\mathrm{e}{-3})$} & {\boldmath $6.56\mathrm{e}{-4}\;(0.0294)$} & $1.93\;(1.13)$ \\
SteinPnP (no ref) & $0.286\;(0.243)$ & $5.77\mathrm{e}{-3}\;(5.12\mathrm{e}{-3})$ & $0.186\;(0.198)$ & $2.08\mathrm{e}{-4}\;(1.56\mathrm{e}{-3})$ & $5.64\mathrm{e}{-3}\;(0.0384)$ & $0.87\;(0.194)$ \\
SteinPnP-Vanilla & $0.5\;(0.459)$ & $8.67\mathrm{e}{-3}\;(9.85\mathrm{e}{-3})$ & $8.88\mathrm{e}{-4}\;(0.0162)$ & $8.87\mathrm{e}{-5}\;(2.09\mathrm{e}{-3})$ & $2.68\mathrm{e}{-3}\;(0.0476)$ & $1.71\;(0.698)$ \\
EPnPU & $0.284\;(0.231)$ & $5.46\mathrm{e}{-3}\;(4.85\mathrm{e}{-3})$ & $0.128\;(0.146)$ & $1.78\mathrm{e}{-4}\;(1.35\mathrm{e}{-3})$ & $4.8\mathrm{e}{-3}\;(0.0377)$ & {\boldmath $0.0113\;(8.15\mathrm{e}{-3})$} \\
DLSU & $0.284\;(0.231)$ & $5.46\mathrm{e}{-3}\;(4.85\mathrm{e}{-3})$ & $0.133\;(0.148)$ & $1.95\mathrm{e}{-4}\;(1.5\mathrm{e}{-3})$ & $5.83\mathrm{e}{-3}\;(0.0364)$ & $0.116\;(0.0469)$ \\
MLPnP & $0.283\;(0.245)$ & {\boldmath $5.4\mathrm{e}{-3}\;(5.14\mathrm{e}{-3})$} & $0.155\;(0.162)$ & $1.5\mathrm{e}{-4}\;(1.31\mathrm{e}{-3})$ & $4.25\mathrm{e}{-3}\;(0.0334)$ & $0.0119\;(0.0428)$ \\
EMCEE (floor) & -- & -- & $-1.11\mathrm{e}{-3}\;(3.14\mathrm{e}{-3})$ & $3.26\mathrm{e}{-6}\;(1.89\mathrm{e}{-4})$ & $9.67\mathrm{e}{-5}\;(3.7\mathrm{e}{-3})$ & $17.4\;(5.42)$ \\
\bottomrule
\end{tabular}}
\vspace*{-4mm}
\end{table*}

Finally, we examine a real sequence exhibiting persistent coplanar ambiguity: a planar AprilTag observed during a camera pitch sweep, with depth increasing linearly from $2\,\mathrm{m}$ at frame~0 to $5\,\mathrm{m}$ at frame~600. Fig.~\ref{fig:apriltag} depicts the pitch marginal distributions of SteinPnP and DLSU frame by frame. A sliced instance at frame 450 is also shown. As the distance to the Apriltag increase, we observe pose ambiguities that are characterized by rapid jumps in pitch angle. DLSU only outputs one mode by design and thus fails to predict the pose ambiguities. On the other hand, we demonstrate that SteinPnP can produce bimodal distributions to capture the pose ambiguities.


\section{conclusion}

In this work, we proposed SteinPnP, a non-parametric solver that quantifies 6-DoF pose uncertainty in the PnP problem. A key advantage of SteinPnP is its ability to flexibly represent both unimodal and multimodal pose uncertainties, which is a phenomenon common to PnP problems where existing Gaussian-based solvers fail. Experimentally, we demonstrate across synthetic and real-world benchmarks that SteinPnP achieves superior uncertainty estimates compared to existing Gaussian solvers. SteinPnP closely matches sampling-based methods while running orders of magnitude faster without compromising accuracy. In the future, we plan to further improve run-time efficiency by exploring parametric models capable of representing complex distributions.

\addtolength{\textheight}{-12cm}   

\bibliographystyle{IEEEtran}
\bibliography{references}

\end{document}